\documentclass[
]{revtex4-2}
\usepackage{booktabs}
\usepackage{graphicx}
\usepackage{dcolumn}
\usepackage{bm}
\usepackage{amsmath}
\usepackage{bbm}
\usepackage{bm}
\usepackage{xcolor}
\usepackage{makecell}
\usepackage{cellspace} 
\usepackage{soul}
\usepackage{amssymb}
\sethlcolor{white}
\usepackage{lineno}

\usepackage{comment}

\begin{document}
\title{ChemKANs for Combustion Chemistry Modeling and Acceleration}

\author{Benjamin C. Koenig}
\thanks{B.C.K. and S.K. contributed equally to this work.}
\affiliation{Department of Mechanical Engineering, Massachusetts Institute of Technology, 77 Massachusetts Ave, Cambridge, MA 02139, United States.}

\author{Suyong Kim}
\thanks{B.C.K. and S.K. contributed equally to this work.}
\affiliation{Department of Mechanical Engineering, Massachusetts Institute of Technology, 77 Massachusetts Ave, Cambridge, MA 02139, United States.}

\author{Sili Deng}
\email{Corresponding author, silideng@mit.edu}
\affiliation{Department of Mechanical Engineering, Massachusetts Institute of Technology, 77 Massachusetts Ave, Cambridge, MA 02139, United States.}
\date{\today}

\begin{abstract}
Efficient chemical kinetic model inference and application in combustion are challenging due to large ODE systems and widely separated time scales. Machine learning techniques have been proposed to streamline these models, though strong nonlinearity and numerical stiffness combined with noisy data sources make their application challenging. Here, we introduce ChemKANs, a novel neural network framework with applications both in model inference and simulation acceleration for combustion chemistry. ChemKAN's novel structure augments the generic Kolmogorov Arnold Network Ordinary Differential Equations (KAN-ODEs) with knowledge of the information flow through the relevant kinetic and thermodynamic laws. This chemistry-specific structure combined with the expressivity and rapid neural scaling of the underlying KAN-ODE algorithm instills in ChemKANs a strong inductive bias, streamlined training, and higher accuracy predictions compared to standard benchmarks, while facilitating parameter sparsity through shared information across all inputs and outputs. In a model inference investigation, we benchmark the robustness of ChemKANs to sparse data containing up to 15$\%$ added noise, and superfluously large network parameterizations. We find that ChemKANs exhibit no overfitting or model degradation in any of these training cases, demonstrating significant resilience to common deep learning failure modes. Next, we find that a remarkably parameter-lean ChemKAN (344 parameters) can accurately represent hydrogen combustion chemistry, providing a 2$\times$ acceleration over the detailed chemistry in a solver that is generalizable to larger-scale turbulent flow simulations. These demonstrations indicate the potential for ChemKANs as robust, expressive, and efficient tools for model inference and simulation acceleration for combustion physics and chemical kinetics. 

\end{abstract}

\maketitle

\section{Introduction}

Chemical kinetic modeling underpins a wide range of scientific and engineering applications, from biological systems to energy conversion. In combustion, two long-standing challenges hinder effective kinetic modeling: the complexity and size of detailed reaction mechanisms \cite{gao_reaction_2016}, and the computational stiffness these systems introduce during numerical integration \cite{lu_toward_2009, kim_stiff_2021}. Recent advances in machine learning offer new opportunities to address both challenges through data-driven model discovery and accelerated surrogate solvers \cite{ihme_combustion_2022}. 

For kinetic model discovery, a variety of learning algorithms, model structures, and optimization approaches have emerged. The Chemical Reaction Neural Network approach \cite{ji_autonomous_2021}, for example, is capable of inferring reaction networks and parameters from limited species trajectory or heat release data \cite{koenig_accommodating_2023, koenig_uncertain_2024, koenig_comprehensive_2025} by directly enforcing the Arrhenius and mass action laws in a neural network structure. The Sparse Identification of Nonlinear Dynamics (SINDy) approach is similarly capable of extracting models from experimental data by assuming various functional relationship building blocks and learning the precise forms needed to fit the data \cite{hoffmann_reactive_2019}. Further optimization and inverse modeling tools exist for other chemical kinetic inference problems \cite{langary_inference_2019, kim_learning_2024, kim2023inference}, with a key piece of many physics-based model inference techniques being a certain (and often substantial) degree of prior knowledge of the governing equations, reaction pathways, and reactants. A critical need across all of these methods is robustness to noisy data and model uncertainty, conditions common in combustion kinetics \cite{ji_autonomous_2021, li_bayesian_2023, koenig_uncertain_2024}.

On the solver front, researchers have proposed methods for dimension reduction~\cite{koenig_accommodating_2023, koenig_multi-target_2023} and computational acceleration~\cite{alqahtani_data-based_2021, jung_hessian-based_2024, owoyele_chemnode_2022, kumar_combustion_2024} to handle stiff, high-dimensional systems. For instance, Owoyele and Pal \cite{owoyele_chemnode_2022} recently proposed ChemNODE, a creative and high-performing tool that uses the neural ODE concept of Chen et al. \cite{chen_neural_2019} to replace a complete chemical kinetic model with a collection of neural networks, one for each tracked thermochemical quantity. By using these networks to directly link the current thermochemical state to the chemical source term with no other problem-specific treatment, computational acceleration was enabled in the as-studied homogeneous reactor while retaining the generalizability of the surrogate model to higher-dimensional reacting flows where such acceleration becomes significantly more meaningful. Other recent works leverage DeepONets \cite{lu_learning_2021} to directly learn stiff integrators using neural operators, either with problem-specific network structures \cite{koenig_kinetic_2023, koenig_multi-target_2023} or by mapping from the current state to the source term, similar to ChemNODE \cite{goswami_learning_2024, kumar_combustion_2024}, allowing for significant computational acceleration downstream.

These strengths all come with drawbacks, however. Owoyele and Pal \cite{owoyele_chemnode_2022}, for example, found that while the neural ODE approach's clever exploitation of the dynamical structure of chemical kinetic models can provide high accuracy, the nonlinearity inherent to such models creates a challenging inference problem for the underlying MLP layers. This led the authors to omit a handful of species (including the key H radical) and break the training up into multiple unique and likely redundant networks, rather than a single cohesive architecture. Similarly, DeepONet techniques are cheap to evaluate and capture steady state behavior well, but their accuracy can suffer in stiff regions of the data that the integrator (which with DeepONets must be inferred directly by the network, as they do not explicitly leverage existing ODE solvers) can learn to skip over without significant penalties. We thus find that despite these recent novel and productive efforts, the training of efficient surrogate models for combustion chemistry remains a challenging task with open questions due to stiff behavior in the solution profiles and numerical instability in the nonlinear training processes.

Kolmogorov-Arnold Networks were proposed recently \cite{liu_kan_2024} as an alternative to multi-layer perceptrons (MLPs) for general neural network applications, where instead of learning weights and biases on fixed activation functions, the shapes and magnitudes of the activation functions themselves are learned via gridded basis function sums and products. This shift was proposed to increase neural convergence rates, accuracy, and generalization. Echoing the development of traditional MLPs, physics-informed KAN structures were proposed shortly after, where it was found that certain knowledge of physical laws embedded in the training process can help the KAN converge to a physically meaningful solution \cite{patra_physics_2024, guo_physics-informed_2024, howard_finite_2024}. Similar developments have been studied where direct encoding of physical insights or specific geometries into novel KAN structures has shown significant promise, such as including physical symmetries for quantum architecture search \cite{kundu2024kanqas} or irregular geometries for flow simulations \cite{kashefi2025kolmogorov}. The inference benefits of KANs were additionally demonstrated to extend to dynamical system modeling in the Kolmorogorov-Arnold Network Ordinary Differential Equations (KAN-ODEs) framework \cite{koenig_kan-odes_2024}, where KANs replaced MLPs in the neural ODE algorithm \cite{chen_neural_2019}, and have since been demonstrated in a variety of settings, including predator-prey dynamics, shock formation, complex equations, phase separation \cite{koenig_kan-odes_2024}, personalized cancer treatment \cite{azimi_kolmogorov-arnold_2025}, and flashover prediction \cite{sezavar_integrating_2025}.

In short, KAN-ODEs leverage KAN networks as gradient getters, while maintaining standard ODE solvers to integrate the solution profiles as in a traditional numerical approach. KAN-ODEs were shown to retain all major KAN benefits while also accessing the dynamical system inference capabilities of the neural ODE framework \cite{koenig_kan-odes_2024}, which would appear to lend them to efficient chemical kinetic system modeling. However, a few key questions remain. KAN-ODEs have so far only been tested in relatively small systems (up to two-dimensional state variables), making their applicability and performance in larger, practical combustion systems unknown. Additionally, KANs in general have been shown to suffer substantially when trained with noisy datasets \cite{shen_reduced_2024}. While we theorize that their direct coupling to ODE integrators combined with their sparse parameterization and smooth activations should provide KAN-ODEs with strong robustness to noise regardless of previously reported issues in generic KANs, this has not yet been studied quantitatively.

In this work, we aim to develop a chemistry KAN-ODE (ChemKAN) framework for chemical kinetic modeling by designing the Kolmogorov-Arnold network gradient getter to learn the \hl{spatially-invariant relationship between the current thermochemical state and the chemical source terms. In the operator splitting regime, such a surrogate can be directly coupled to existing CFD or machine learning-based flow solvers for multi-dimensional combustion simulations in arbitrary physical domains}. Developed across two case studies of increasing complexity, the ChemKAN framework contains a physics-informed, two-stage training process that enforces the direct coupling between species production and heat release, and additionally contains a soft constraint for element conservation. We further stabilize the optimization problems by implementing forward sensitivity analysis. Stiff chemistry is fully resolved via an attached numerical ODE solver. We study two cases here to explicitly probe the key behaviors and gaps in the model inference and solver acceleration literature identified above, both of which are addressed with ChemKANs. 

First, we demonstrate the capability of ChemKANs to extract realistic and multi-species models from synthesized experimental data in a comparison against DeepONets (DONs) in biodiesel production modeling. In this case, increasing levels of noise in the training data test the abilities of the two different approaches to extract the true underlying behavior, and evaluate the robustness of ChemKANs (and KAN-ODEs in general) to noisy data in light of the recent work suggesting the weakness of KANs in the presence of noise \cite{shen_reduced_2024}. Second, we demonstrate ChemKANs as efficient and time-saving surrogate models in an even larger system by learning zero-dimensional hydrogen combustion behavior using homogeneous reactor data that are subject to stiff dynamics, in a study designed to facilitate direct comparison against the MLP-based ChemNODE structure \cite{owoyele_chemnode_2022} that this second ChemKAN application was inspired by. In contrast to ChemNODE, where a reduced subset of the thermochemical state (excluding H, HO$_2$, and H$_2$O$_2$) was learned using separately trained, non-interacting networks, we learn all species and temperature profiles here with a single compact ChemKAN network while retaining similar computational acceleration and performance. Across these two distinct cases, we demonstrate the strong capability and robustness of ChemKANs as efficient and expressive tools for both modeling and inference in combustion chemistry.

\section{Methods}

This section describes the mathematical details in the newly developed chemistry Kolmogorov-Arnold network ordinary differential equations (ChemKANs). We begin with a review of a zero-dimensional chemical kinetic model without transport effects in Sec.~\ref{methods:background}, followed by existing MLP-based kinetic modeling techniques in Sec.~\ref{methods:prevailing}. Then, we discuss the implementation of the ChemKAN framework for multi-purpose chemical model inference and computational acceleration in Sec.~\ref{methods:chemkan_odes}. That subsection includes a physics-enforced ChemKAN architecture, optimal learning strategies, and physics-informed loss functions. Finally, in Sec.~\ref{methods:data_gen} we provide the kinetic models of biodiesel pyrolysis and hydrogen-air combustion that are used to demonstrate the performance of ChemKANs.

\subsection{Chemical kinetics model}\label{methods:background}

Given the state variables $\mathbf{u}(t)=[Y_{1}, Y_{2}, ..., Y_{m}, T](t)$ where $T$ is the temperature, $Y$ is the mass fraction, and $m$ is the number of species, the net production/consumption rate for each species in a homogeneous reactor can be expressed as

\begin{equation} \label{eq:species}
    \frac{d Y_{i}}{d t} = \frac{1}{\rho}  W_{i} \dot{\omega}_{i},
\end{equation}

\noindent where $t$ is the time, $\rho$ is the density, $W_{i}$ is the molecular weight of species i, and $\dot{\omega}$ is the molar production or consumption rate \cite{kee2005chemically}. In the commonly used operator splitting approach, this homogeneous reactor is also applicable to higher-dimensional, turbulent simulations. While the specific functional form may differ across systems, $\dot{\omega}$ is typically a strong function of temperature (for example, in the Arrhenius form $\dot{\omega} \propto \exp({-E_a/RT})$) as well as the current species concentrations.  Additionally, energy conservation can be modeled by tracking the system temperature as in Eq.~\ref{eq:energy}.

In some cases, Eq. \ref{eq:species} sufficiently describes a chemical process when heat release or consumption is negligible. In other cases, such as combustion and pyrolysis processes, chemical reactions entail exothermic and endothermic behaviors. In these cases, the temperature of a system can be tracked with energy conservation, as in 

\begin{equation} \label{eq:energy}
    \frac{d T}{d t} = - \sum_{i=1}^{m}  \frac{h_{i} \dot{Y}_{i}}{c_{p}},
\end{equation}

\noindent where $c_{p}$ is the mixture-averaged specific heat and $h_{i}$ is the enthalpy of species $i$. Here, the strongly nonlinear coupling of Y and T in Eqs. \ref{eq:species} and \ref{eq:energy} often leads to modeling challenges such as numerical stiffness. While always present in the governing laws and essential to accurate modeling for processes such as combustion and pyrolysis, Eq. \ref{eq:energy} can occasionally be neglected, such as when modeling the kinetic rates of an isothermal experiment. We can express Eqs.~\ref{eq:species}-\ref{eq:energy} as a generic system of equations  $\mathbf{f}$ such that

\begin{equation} \label{eq:dynamical_system}
    \frac{d\mathbf{u}}{dt} = \mathbf{f}\left(\mathbf{u}, t \right).
\end{equation}

\noindent Therefore, the thermochemical states $\mathbf{u}$ can be predicted by integrating $\mathbf{f}\left(\mathbf{u}, t \right)$ with an ODE integrator over time.

\subsection{MLP-based models} \label{methods:prevailing}

Recent machine learning approaches for model inference and computational acceleration often rely on neural networks constructed from Multi-layer perceptrons (MLPs). Among them, we introduce two mainstream models based on deep operator networks and neural ordinary differential equations. We will use these two MLP-based models to highlight the performance of our ChemKANs (to be introduced in Sec.~\ref{methods:chemkan_odes}).

\subsubsection{Deep Operator Network (DeepONet)}

As one of the more popular architectures for combustion applications \cite{koenig_kinetic_2023, koenig_multi-target_2023, goswami_learning_2024, kumar_combustion_2024}, DeepONets \cite{lu_learning_2021} learn the chemical kinetic system (outlined in Sec. \ref{methods:background}) through a physics-inspired separation between the system's parameterization and the solution coordinate (Fig.~\ref{fig:comparison}(A)). Specifically, the solution is learned using two neural networks (branch net for thermochemical states $\mathbf{u}$ and trunk net for time $t$), as per 

\begin{equation} \label{eq:DON_formulation}
    \mathbf{u}(\mathbf{u}(0),t) = \text{MLP}_{\text{opt}}[\text{MLP}_{\text{br}}(\mathbf{u}(0), \bm{\theta}_{\text{br}}) \odot \text{MLP}_{\text{tr}}(t, \bm{\theta_{\text{tr}}}), \bm{\theta}_{\text{opt}}].
\end{equation}

\noindent More specifically, given an initial condition $\mathbf{u}(0)=[\mathbf{Y}(t=0), T(t=0)]$, the DeepONet reports the solution $\mathbf{u}$ at a given time $t$ as the element-wise product ($\odot$) of the branch network MLP$_{\text{br}}$ evaluated on the initial condition and the trunk network MLP$_{\text{tr}}$ evaluated on the current time, with an optional final MLP$_{\text{opt}}$ layer for additional nonlinear encoding. $\bm{\theta}$ are the learnable parameters for the respective neural networks. The DeepONet here, as well as in other augmented structures from higher-complexity implementations \cite{goswami_learning_2024, kumar_combustion_2024}, directly learns the solution state at future times. By eliminating the integration step, these methods have shown substantial acceleration of computational times.

\subsubsection{ChemNODEs}

In contrast to the DeepONet approach, ChemNODE \cite{owoyele_chemnode_2022} aims to learn the source terms $\mathbf{f}$ in Eq.~\ref{eq:dynamical_system}, rather than the direct thermochemical states over time $\mathbf{u}(t)$. However, training challenges were observed \cite{owoyele_chemnode_2022} when testing a single MLP network with $m+1$ inputs and $m+1$ outputs ($m$ for $\mathbf{Y}$ and 1 for $T$). This led to the development of a segregated model for each thermochemical state $u_{i}$, as per

\begin{equation} \label{eq:chemnode}
    \frac{d{u}_{i}}{dt}(t) = \text{MLP}_{i}\left(\mathbf{u}\left(t\right), \bm{\theta}_{i}\right).
\end{equation}

\noindent Therefore, ChemNODE constructs $m+1$ MLP networks, each of which reads the entire current thermochemical state vector $\mathbf{u}$. The $i^{\text{th}}$ network has its own unique set of parameters $\bm{\theta}_i$, and is responsible for computing the current temporal gradient of that scalar $u_i$. Notably, ChemNODE is still trained via MSE loss computed against the integrated solution profile (rather than the gradients that the MLPs directly output), thanks to the differentiable Neural ODE framework \cite{chen_neural_2019}. This latent-dynamics modeling for $\mathbf{f}$ allows ChemNODEs to compute rate terms without restriction to a specific instance. This contrasts with standard DeepONets, which require an initial condition fed into the branch network. Therefore, ChemNODEs can be effortlessly generalized as the interpretable source terms even for large-scale simulations.

\begin{figure*}[tb]
    \centering
    \includegraphics[width=0.95\linewidth]{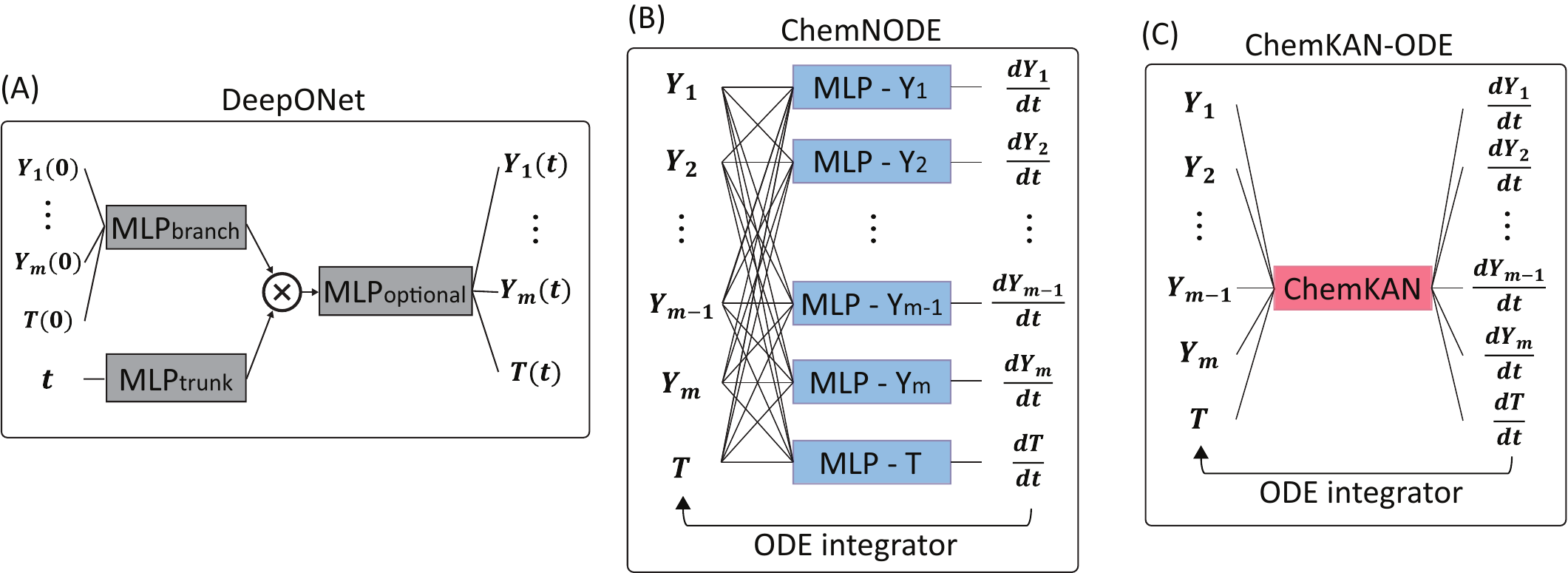}
	\caption{Comparison of three kinetic modeling approaches discussed in this work. (A) DeepONet, where the initial $(m+1)$-dimensional state and current time are input to two inductively split networks (and a third optional network) to output the current state. (B) ChemNODE, where the complete current state is input in parallel through $m+1$ separate networks, each of which outputs a single state gradient, the collection of which is integrated through a differentiable ODE solver to reach the next time step. (C) ChemKAN, which has a similar workflow to ChemNODE but replaces the $m+1$ MLP networks with a single, physics-mimicking KAN-based structure that is capable of computing the entire state gradient in a single pass. }
    \label{fig:comparison}
\end{figure*}

We remark again that the split networks $\text{MLP}_{i}$ in this framework appear inefficient. In fact, this multi-network approach requires an $(m+1)$-step training process, where each network is trained with the remaining thermochemical scalars held frozen. While this separated training strategy facilitates model convergence (and in fact was found necessary~\cite{owoyele_chemnode_2022} to converge the MLP gradient getters), it increases training cost significantly with the number of species, especially when considering that no knowledge is shared between networks, even for common reactants. See Fig. \ref{fig:comparison} for visualizations of these two existing methods, as well as a simplified depiction of the method we will propose in the next sections.

\subsection{Chemistry Kolmogorov-Arnold network ordinary differential equations (ChemKANs)} \label{methods:chemkan_odes}

\subsubsection{Vanilla KAN-ODEs} \label{methods:vanilla_kan}

The KAN-ODEs framework was proposed by Koenig et al. \cite{koenig_kan-odes_2024} to model a dynamical system in the form of differential equations, where the gradient function is replaced by a KAN network of $L$ layers,

\begin{equation} \label{eq:kan-odes}
    \frac{d\mathbf{u}}{dt} = \text{KAN}\left(\mathbf{u}\left(t\right), \bm{\theta}\right) =
    \left({\Psi}_{L-1}\circ {\Psi}_{L-2}\circ\cdots\circ {\Psi}_{1}\circ {\Psi}_{0}\right)\left(\mathbf{u\left(t\right)}\right),
\end{equation}

\noindent where KAN is the KAN representation of the system equation, parameterized by $\bm{\theta}$. The original KAN structure \cite{liu_kan_2024}, so-called AddKAN, connects an $n_l$-sized input to an $n_{l+1}$-sized output with the learnable activation function matrix such that

\begin{equation}
\begin{aligned}\label{eq:kanforwardmatrix}
     \text{AddKAN: }{\Psi}_{l}^{\text{add}} = {\Phi}_l  
     =\begin{pmatrix}
        \phi_{l,1,1}(\cdot) & \phi_{l,1,2}(\cdot) & \cdots & \phi_{l,1,n_{l}}(\cdot) \\
        \phi_{l,2,1}(\cdot) & \phi_{l,2,2}(\cdot) & \cdots & \phi_{l,2,n_{l}}(\cdot) \\
        \vdots & \vdots & & \vdots \\
        \phi_{l,n_{l+1},1}(\cdot) & \phi_{l,n_{l+1},2}(\cdot) & \cdots & \phi_{l,n_{l+1},n_{l}}(\cdot) \\
    \end{pmatrix},
\end{aligned}
\end{equation}

\noindent where each $\phi$ is a unique learnable activation function connecting a single input to a single output (thus ${\Phi}_l \in \mathbb{R}^{n_l \times n_{l+1}}$). In other words, each input is connected to each output with a unique learnable activation function (much like in an MLP, where each input is connected to each output with a unique learnable weight), leading to a total of $n_{l} \cdot n_{l+1}$ activation functions connecting the $l^{\text{th}}$ and $(l+1)^{\text{th}}$ layers. We use RBF basis functions in the current work as was shown previously for KANs \cite{li_kolmogorov-arnold_2024} and KAN-ODEs \cite{koenig_kan-odes_2024, koenig_leankan_2025}, although the choice of $\phi$ is is flexible, and many other options have been proposed in the literature including B-splines \cite{liu_kan_2024}, ReLU functions \cite{qiu_relu-kan_2024}, and various other combinations \cite{ta_af-kan_2025, ta_fc-kan_2025}.  The AddKAN structure has an inherent problem of expression using only additive operations, limiting its concise expressivity for problems involving substantial use of the multiplication operator. Therefore, recent studies have proposed new layer structures to address this issue and improve parameter efficiency \cite{liu_kan_2024_2,koenig_leankan_2025}. Here we use LeanKAN \cite{koenig_leankan_2025}, which has shown promise to be the most effective and efficient for both additive and multiplicative operations. The LeanKAN structure is achieved by summation of two separate terms, $\mathbf{y}_{l}^{\text{add}}$ and $\mathbf{y}_{l}^{\text{mult}}$, such that

\begin{equation} \label{eq:new_output}
    \text{LeanKAN: }\Psi_{l}^{\text{lean}}(\mathbf{x}_l)=\mathbf{y}^{\text{mult}}_{l} + \mathbf{y}^{\text{add}}_{l} \in \mathbb{R}^{n_{l+1}},
\end{equation}

\begin{align} 
    {y}_{l,i}^{\text{mult}}&= \prod_{j=1}^{n_l^{mu}} {\phi_{l,i,j}}\left(x_{l,j}\right)~~~~~~~~~~~\text{for}~i\in\{1,2, ..., n_{l+1}\} \subset \mathbb{N},\label{eq:rowprod} \\ 
    {y}_{l,i}^{\text{add}}&=\sum_{j=n_l^{mu}+1}^{n_{l}}{\phi_{l,i,j}}\left(x_{l,j}\right)~~~~~\text{for}~i\in\{1,2, ..., n_{l+1}\} \subset \mathbb{N},\label{eq:rowsum}
\end{align}

\noindent where $n_{l}^{mu}$ is the multiplication hyperparameter that dictates the number of multiplication input nodes for layer $l$, and $\phi$ are the same univariate activation functions used in AddKAN (to be defined below). Overall, the LeanKAN formulation takes the $n_l \times n_{l+1}$ matrix of activated inputs, computes sums and products in parallel according to $n^{mu}$ to construct the intermediate $\mathbf{y_l}$, and finally sums the multiplication and addition components of $\mathbf{y_l}$ to arrive at the next layer. Further derivation and implementation details are available for KANs \cite{liu_kan_2024}, KAN-ODEs \cite{koenig_kan-odes_2024}, and LeanKAN layers \cite{koenig_leankan_2025}.

Finally, the activation functions themselves can be expressed with gridded basis functions \cite{li_kolmogorov-arnold_2024} as per 

\begin{align}
    \phi_{l, \alpha, \beta} \left(\text{x} \right) &= \sum_{i=1}^{N} w^{\psi}_{l, \alpha, \beta,i} \cdot \psi \left( \lvert \lvert \text{x}-c_{i} \rvert \rvert \right) + w^b_{l, \alpha, \beta} \cdot b\left(\text{x}\right),\label{eq:basis}\\
    \psi(r)&=\exp(-\frac{r^{2}}{2h^{2}}), \label{eq:RBF}
\end{align}

\noindent where we define $N$ as the grid size, or number of superimposed basis functions used to construct a single activation. $w^{\psi}_{l, \alpha, \beta,i}$ and $w^b_{l, \alpha, \beta}$ are the learnable network parameters that make up $\bm{\theta}$, where $w^{\psi}_{l, \alpha, \beta,i}$ scales each gridpoint's RBF basis functions $\psi(r)$ within the sum, and $w^b_{l, \alpha, \beta}$ scales a single base activation function $b(\text{x})$. $\alpha$ and $\beta$ denote the input-output pair for which the current activation function is defined (e.g., the activation function with subscripts $\alpha = 3$ and $\beta = 2$ connects the second input node to the third output node). The grid itself is defined by its individual gridpoints $c_i$ and gridpoint spacing (or RBF spreading parameter) $h$. While Liu et al. \cite{liu_kan_2024} suggest periodically re-gridding these basis functions to ensure they are learning on the proper input range, here we instead normalize at each layer input using the hyperbolic tangent function as in more recent works \cite{blealtan_efficient-kan_2024, puri_kolmogorovarnoldjl_2024, koenig_kan-odes_2024, koenig_leankan_2025} for computational efficiency. The single base activation term $b(\text{x})$ in each $\phi$ is a Swish activation function \cite{ramachandran_searching_2017}.

\subsubsection{Novel chemistry KAN (ChemKAN) architecture}\label{methods_newkan}

Here, we design a novel ChemKAN architecture through a unique composition of AddKAN and LeanKAN layers that shows invariance to the number of species by combining all model behavior into a single network architecture (see Fig.~\ref{fig:comparison}). The standard KAN-ODE architecture has the same dimensions of inputs and outputs ($m+1$ thermochemical states in our current problem). In contrast, the ChemKAN architecture mimics the structure of the actual governing equations in Eqs.~\ref{eq:species}$\sim$\ref{eq:energy} to redirect the kinetic and thermal inputs and outputs in a physics-inspired manner. We define the full and species-only state variables with $\mathbf{u} = [\tilde{\mathbf{u}}, T]$, and separate the kinetic and energy equations such that

\begin{align} 
    \frac{d\tilde{\mathbf{u}}}{dt} &= \text{KAN}_{\text{kin}}\left(\mathbf{u},\bm{\theta}_{\text{kin}} \right), \label{eq:chemkan_species}\\
    \frac{d {T}}{d t} &= \text{Linear}\left(\frac{d\tilde{\mathbf{u}}}{dt},{\bm{\theta}}_{\text{thermo}} \right) + \epsilon \label{eq:chemkan_energy} \\
    &= \text{Linear}\left(\text{KAN}_{\text{kin}}\left(\mathbf{u},\bm{\theta}_\text{kin} \right),{\bm{\theta}}_{\text{thermo}} \right) + \text{KAN}_{\text{cor}}(\mathbf{u}, {\bm{\theta}}_{\text{cor}}). \label{eq:chemkan_energy_cor}
\end{align}

\begin{figure*}[tb]
    \centering
    \includegraphics[width=0.85\linewidth]{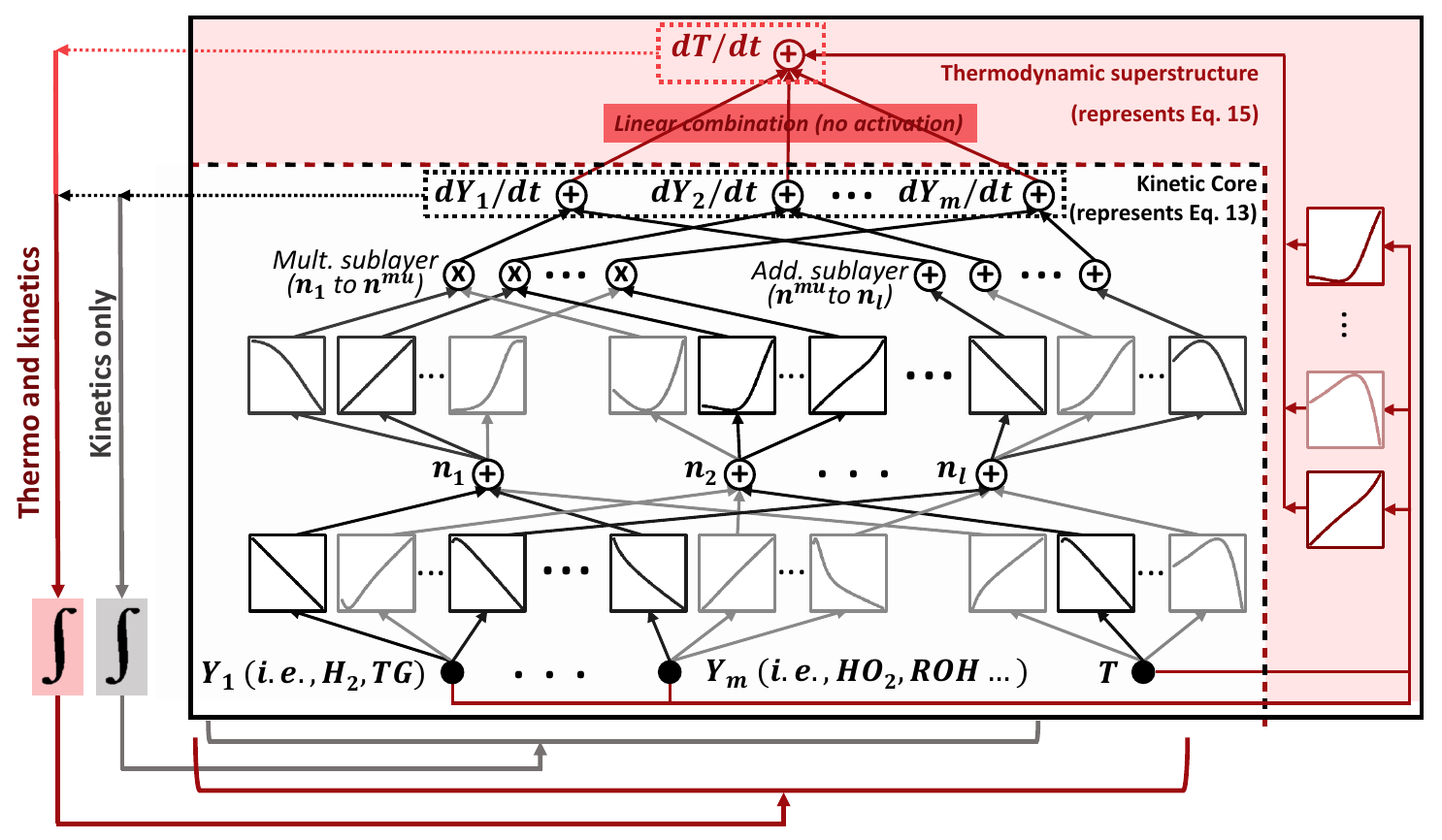}
	\caption{Proposed ChemKAN structure to embed physical knowledge in KAN-ODEs for chemical kinetic modeling. The grey portion of the network contains $m+1$ inputs (all species and temperature) and $m$ outputs (all species rates), and represents the kinetic behavior of the chemical model. The red portion of the network performs a linear transformation on the species rates, with a nonlinear thermophysical parameter correction, to account for the thermodynamic behavior leading to heat release and temperature change. The ChemKAN can either be trained and evalauted for kinetics only (as in the biodiesel case of Secs. \ref{methods:bio} and \ref{results:bio}), or trained sequentially for kinetics and then thermodynamics, with both portions of the network evaluated in serial for applications predicting both kinetic and thermodynamic behavior (as in the H$_2$ case of Secs. \ref{methods:h2} and \ref{results:H2}).}
    \label{fig:kanode}
\end{figure*}

\noindent We define the terms in these equations line by line throughout this paragraph. In Eq. \ref{eq:chemkan_species}, we encode prior knowledge of the true species production/consumption rate relationship of Eq. \ref{eq:species} into the ChemKAN by computing the species-only production rate ${d\tilde{\mathbf{u}}}/{dt}$ from the entire state input $\mathbf{u}$, via a KAN network (KAN$_\text{kin}$) parameterized by $\bm{\theta}_{\text{kin}}$. Then, from the true energy equation (Eq. \ref{eq:energy}), we recognize that the temperature rate $dT/dt$ is a simple linear combination of the species production rates, with scaling factors defined by the enthalpy $h$ and specific heat values $c_p$ (or alternatively, in Eq. \ref{eq:chemkan_energy} by the $m$ scalars in the linear mapping parameterized by $\bm{\theta}_{\text{thermo}}$). Thus, the crux of the thermodynamic superstructure is a computationally trivial, simple linear sum of the already-evaluated outputs of the kinetic core, as shown in Eqs. \ref{eq:chemkan_energy} and \ref{eq:chemkan_energy_cor}. One level deeper, we recognize that a secondary effect in the true Eq.~\ref{eq:energy} is the dependence of the thermophysical parameters, specifically $c_p$, on the temperature and species mixture. The error stemming from the first-order Linear($\cdot$) approximation's failure to account for such thermophysical parameter variation is reflected in the $\epsilon$ term of Eq. \ref{eq:chemkan_energy}, which in the final formulation of Eq. \ref{eq:chemkan_energy_cor} is accounted for via a supplemental single-layer, single-output KAN correction carrying forward a functional dependence on the species and temperature inputs, parameterized by $\bm{\theta}_{\text{cor}}$. Overall, ChemKAN is composed of a kinetic core structure and a thermodynamic superstructure that strongly mimic the true governing equations, operate largely in series, and include full sharing of all reaction and species production information. This architecture allows for versatile and flexible modeling by turning ``on'' or ``off'' the energy equation for standalone, kinetic core-only modeling or combined kinetic and thermodynamic modeling.

\subsubsection{Network structure details}\label{methods:chemkanode}

The ChemKAN structure proposed here is combined with the KAN-ODE framework detailed in Sec. \ref{methods:vanilla_kan} to leverage both the physics-guided ChemKAN architecture and the proven dynamical system benefits offered by the KAN-ODE integration. For nomenclature consistency, we henceforth refer to both the newly designed network structure as well as its complete integration with an ODE solver as ChemKAN. For the relatively larger kinetic core responsible for learning all reaction information, we define the internal layer structure as

\begin{equation} \label{eq:kan-odes-our-study}
    \text{KAN}_{\text{kin}}\left(\mathbf{u}\left(t\right), \bm{\theta}_{\text{kin}}\right) =
    \left({\Psi}_{1}^{\text{lean}}\circ  {\Psi}_{0}^{\text{add}}\right)\left(\mathbf{u}\left(t \right)\right),
\end{equation}

\noindent a two-layer structure per the notation introduced in Eq. \ref{eq:kan-odes}. The first is an AddKAN layer \cite{liu_kan_2024} (which is equivalently \cite{koenig_leankan_2025} a LeanKAN layer with $n^{mu}=0$). The second is a LeanKAN layer with $n^{mu}>0$ to inject the multiplication operator. Further discussion of this specific two-layer form and its merits is available separately~\cite{koenig_leankan_2025}. The thermodynamic correction structure simply comprises a scalar sum of univariate activations evaluated on the $m+1$ components of $\mathbf{u}$,

\begin{equation}
    \text{KAN}_{\text{cor}}\left(\mathbf{u}\left(t\right), {\bm{\theta}}_{\text{cor}}\right)= \Psi_{0}^{\text{add}}\left( \mathbf{u}\left(t \right) \right).  
\end{equation}

\noindent As in ChemNODE, differentiable ODE solvers are leveraged to enable ChemKAN encodings of the state gradients to be trained on the integrated state profiles.

\subsubsection{Loss function} \label{methods:loss}

We define the loss function used for ChemKAN training as
\begin{equation}
\begin{aligned}
    \mathcal{L}\left(\bm{\theta} \right) &= \mathcal{L}_{\text{MSE}}\left(\bm{\theta} \right) + \mathcal{L}_{\text{PINN}}\left(\bm{\theta} \right) \\&= \underbrace{\frac{1}{n^*}  \sum_{j=1}^{N_{t}}\sum_{k=1}^{n^*} {\left(\hat{u}^{\text{pred}}_{k}\left( t_{j}, \bm{\theta}\right)- \hat{u}^{\text{obs}}_{k}\left(t_{j}\right)\right) ^{2}}}_{\text{MSE, variable }n^*} + \underbrace{\alpha_{\text{PINN}} \sum_{i=1}^{N_e} \sum^{N_t}_{j=1} \left\lvert  \sum^{m}_{k=1} \frac{N_i^kW_i(Y_{k, j}^{\text{pred}}-{Y}^{\text{pred}}_{k, 1})}{W_k} \right\rvert} _{\text{optional element conservation}}, \label{eq:loss_h2} \\
    & \text{where } n^{\ast} = \left\{
    \begin{array}{ll}
      m, & \text{in Stage 1.}\\
      m+1, & \text{in Stage 2.}
    \end{array}
  \right.
  \end{aligned}
\end{equation}

\noindent Here ${\hat{u}_k}$ denotes the $k^{th}$ thermochemical state normalized to the [0, 1] window by subtracting the minimum then dividing by the range. $\hat{u}_k^{\text{pred}}(t_j, \bm{\theta})$ is the network prediction for this state quantity at time $t_j$ with the network parameters $\bm{\theta}$ (including the kinetic, thermo, and correction parameters), while $\hat{u}^{\text{obs}}_k(t_j)$ is the corresponding training data. $N_t$ is the number of datapoints in the temporal profiles. In the first MSE term, $n^*=m$ when training the kinetic core, as only species profiles are learned. For the thermodynamic superstructure, $n^*=m+1$ is used to train the added temperature output. We also provide an optional element conservation physics-informed loss term, or PINN term \cite{raissi_physics-informed_2019,kumar_physics-informed_2024}, to encourage the ChemKAN to find models that obey physical laws. There, $N_e$ is the number of elements in the data (i.e., H, O, N). For a given element $i$, the element conservation term begins by computing the mass fraction difference across all $m$ species between the current ChemKAN timestep and the initial condition. This is then converted to an elemental conservation difference via $N_i^k$, the atom count of element $i$ in species $k$; $W_i$, the atomic mass of element $i$; and $W_k$, the molar mass of species $k$. This conserved term is computed at all timesteps $j$ across all elements $i$, and then weighted by $\alpha_{\text{PINN}}$ (here $\alpha_{\text{PINN}}=10^{-4}$). In the examples below, we use only the MSE loss for all biodiesel model inference results, and the MSE loss with the PINN term for H$_{2}$ model acceleration.

\subsubsection{Training Process} \label{methods_training}

We highlight again that this architecture incorporates significant inductive bias by separating the inference of the kinetic behavior and thermodynamic behavior into distinct layers and activations within the complete structure, giving the network an explicit functional coupling between the temperature rate and the species rates to ease its training burden while retaining an accurate model of the real exothermic and endothermic behavior of the system. In itself, this serially run inductive split should theoretically ease some of the nonlinear training difficulties observed with ChemNODE \cite{owoyele_chemnode_2022} while retaining a single cohesive and information-sharing structure. Additionally, and more importantly to stable training and convergence behavior, it allows us to separate the training itself into a kinetic and a thermodynamic stage to greatly reduce the inference burden, as discussed here:

\begin{itemize}
    \item \textbf{Training stage 1---\textit{core kinetics} $\bm{\theta}_\text{kin}$:}  All $m+1$ inputs are used, and only the $m$ species production rate outputs are learned (see Eq. \ref{eq:chemkan_species}). The thermodynamic superstructure of Eqs.~\ref{eq:chemkan_energy} and \ref{eq:chemkan_energy_cor} is not used in this stage, and the input temperatures are simply read in from the training data to provide the kinetic core of the network with a simpler training task. See the grey highlight in Fig. \ref{fig:kanode}. For cases without heat release, this step in isolation is sufficient for a complete model. With heat release, we move to stage 2 once stage 1 is converged.
    \item \textbf{Training stage 2---\textit{thermodynamic superstructure} $\bm{\theta}_\text{thermo}$ \textit{and} $\bm{\theta}_\text{cor}$:}  Once stage 1 is converged, the thermodynamic superstructure is added and the temperature rate is explicitly learned. The entire network (Eq. \ref{eq:chemkan_species} stacked with Eq. \ref{eq:chemkan_energy_cor}) is updated in order to infer the temperature together with all species. See the red highlight in Fig. \ref{fig:kanode}, which is stabilized during training with the already-converged behavior of the grey kinetic core.
\end{itemize}

This two-stage training process contrasts with the $m+1$ stage training process of ChemNODE \cite{owoyele_chemnode_2022}, which requires $m+1$ networks to all learn their own distinct representations of what we know to be shared kinetic and thermodynamic governing laws. The current ChemKAN approach, while trained in two distinct stages, has an overall structure resembling a simple feedforward KAN thanks to its stacked design, where all kinetic and thermodynamic information is shared across all inputs and outputs. For downstream evaluation, a single forward pass through the combined network of Fig.~\ref{fig:kanode}, or alternatively through Eqs.~\ref{eq:chemkan_species}$\sim$\ref{eq:chemkan_energy_cor}, predicts the complete thermochemical state vector $\mathbf{u}(t)$.

While adjoint sensitivity analysis was used in the original KAN-ODE paper \cite{koenig_kan-odes_2024} to compute the gradients of a loss function $d\mathcal{L}/d\bm{\theta}$, in the current work we note the numerical instability present in many chemical kinetic modeling problems due to numerical stiffness \cite{ji_stiff-pinn_2021, goswami_learning_2024, kumar_combustion_2024, owoyele_chemnode_2022, kim_stiff_2021}, potentially leading to failures with adjoint sensitivity analysis in the training process \cite{kim_stiff_2021}. To prevent this, we implement forward sensitivity analysis to mitigate this potential stiffness issue in the ODE solver \cite{kim_stiff_2021}. ChemKANs and their corresponding forward sensitivity equations are solved using an ODE integrator of Tsit5 (Tsitouras 5/4 Runge-Kutta method \cite{tsitouras_rungekutta_2011}). The learnable parameters $\bm{\theta}$ are trained by the Adam optimizer \cite{kingma_adam_2017} with a learning rate of $2 \times 10^{-3}$.

\subsection{Data generation for case studies} \label{methods:data_gen}

We introduce two chemical reaction systems to demonstrate ChemKANs as a multi-purpose modeling technique. The first case of biodiesel production will illustrate the effectiveness of ChemKANs for inferring a chemical kinetic model purely from noisy data. Then, we show computational acceleration for high-fidelity simulations using ChemKANs in a hydrogen-air combustion example subject to significant numerical stiffness.

\subsubsection{Model inference from noisy data -- biodiesel production} \label{methods:bio}

Biodiesel production involves the transesterification of branched triglyceride molecules (TG) with methanol into straight-chain methyl ester molecules, described in Ref.~\cite{darnoko_kinetics_2000} and previously modeled in Refs.~\cite{ji_autonomous_2021, burnham_inference_2008}. A motivating goal here is to evaluate the ChemKAN framework's capability to learn the true models underlying experimental data in a preliminary and numerically well-behaved case related to combustion chemistry, and to probe its robustness to noisy data masking the true underlying signal.

With the three byproducts di-glyceride (DG), mono-glyceride (MG), and glycerol (GL), the three-reaction system can be expressed as 
\begin{align}
TG + ROH {}&\xrightarrow[]{k_1} DG + R'CO_2R,
\\
DG + ROH {}&\xrightarrow[]{k_2} MG + R'CO_2R,
\\
MG + ROH {}&\xrightarrow[]{k_3} GL + R'CO_2R,
\end{align}

\noindent where the reaction rates scale with temperature via the standard Arrhenius law as $k_i = A_{i}\exp(-E_{a, i}/RT)$, with $i=3$ for the three reactions. As in Ref.~\cite{ji_autonomous_2021}, we generate data in this case using $E_a=[14.54, 6.47, 14.42]$ kcal/mol and $\text{ln}(A)=[18.60, 7.93, 19.13]$, with isothermal experiments at temperatures randomly sampled in the range of $323$K to $343$K. We define the species scalar quantities $\mathbf{Y}$ here as concentrations rather than mass fractions, to match the convention with the governing equations. Initial TG and ROH concentrations are randomly sampled uniformly between $0.5$ and $2$ with all other intermediate and output species initialized at zero. 20 training data sets and 10 testing data sets are generated, with 30-second time windows in both consisting of $30$ sampled points. The temperature-dependent yet isothermal reaction rates present in this system motivate the use of the kinetic ChemKAN core structure only (see the Hydrogen example below for use of the kinetic core together with the thermodynamic superstructure).

In this case we additionally probe the effectiveness of ChemKANs in the presence of significant experimental noise. To do so, we add increasing amounts of noise to the data and evaluate ChemKAN's performance against that of a standard Deep Operator Network (DeepONet) \cite{lu_learning_2021}.

\subsubsection{Model acceleration -- hydrogen-air combustion}\label{methods:h2}

Despite its relatively simple chemistry, hydrogen-air combustion is subject to numerical stiffness, leading to high computational costs and numerical instability in the ODE solver. 
The main goal of the ChemKAN here is to achieve high accuracy in species and temperature profile reconstruction while reducing computational cost compared to the detailed chemistry solver. 
We generate training data in Cantera \cite{cantera}, using the H$_{2}$/O$_{2}$ mechanism from GRI-Mech 3.0 \cite{GRImech3} (9 species, 29 reactions). Data are generated at all 36 combinations of initial temperatures $T_0$ in \{950, 1000, 1050, 1100, 1150, 1200\} K and equivalence ratios $\Phi$ in \{0.7, 0.9, 1.1, 1.3, 1.5\}, with the [1150 K, 1.3] case withheld as unseen testing data. The problem setup here is largely identical to that studied originally in ChemNODE \cite{owoyele_chemnode_2022} to facilitate fair comparison, although here we add the single withheld testing dataset to probe ChemKAN's robustness (thus 35 training datasets are used, compared to the 36 in ChemNODE).

In addition to significantly increased numerical stiffness and complexity, a key fundamental difference between the current combustion system and the previous biodiesel synthesis case is the presence of substantial two-way temperature coupling as per Eqs.~\ref{eq:species}$\sim$\ref{eq:energy}. To account for this, we include the thermodynamic ChemKAN superstructure and the two-stage training process outlined in Secs.~\ref{methods_newkan}$\sim$\ref{methods_training}.

\section{Results and Discussion}

\subsection{Biodiesel model inference from noisy data} \label{results:bio}

We begin with an analysis of the biodiesel model inference of the ChemKAN (the kinetic core only for this isothermal case), and a comparison against a traditional DeepONet for an identical task. DeepONets were selected as the target of comparison due to their general use in the scientific machine learning community, to provide a baseline against which to evaluate our first ChemKAN results. Note that comparison against ChemNODE is saved for the hydrogen-air combustion case in Sec.~\ref{results:H2}.

\subsubsection{Model performance of the proposed ChemKAN architecture}

\begin{figure*}[htb!]
    \centering
    \includegraphics[width=0.75\linewidth]{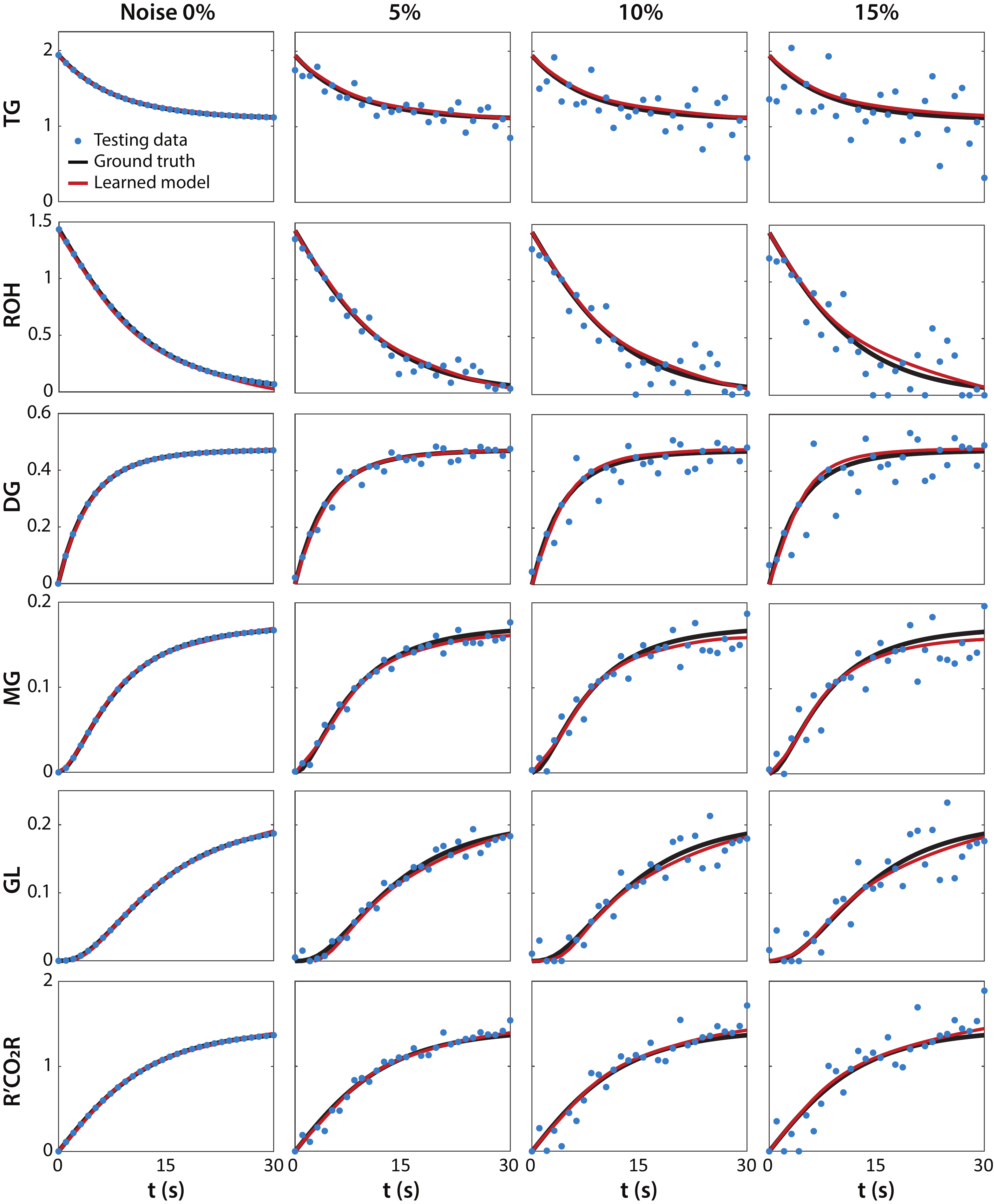}
	\caption{Ground truth and prediction by the learned ChemKAN model for an unseen test case. Each column shows the evolution of species concentrations trained at different noise levels from 0\% to 15\% given the unseen initial condition of [TG] = 1.94, [ROH] = 1.43, [DG] = 0.0, [MG] = 0.0, [GL] = 0.0, [R'CO$_{2}$R] = 0.0, and $T$ = 334.8 K. Noise is shown in the unseen test data to visualize the effect of the noise on the seen training data.}
    \label{fig:performance}
\end{figure*}

The kinetic core (Eq.~\ref{eq:chemkan_species}) of the ChemKAN comprises two layers: one AddKAN and one LeanKAN, as discussed in Sec.~\ref{methods_newkan}. Note that the thermodynamic superstructure (Eq.~\ref{eq:chemkan_energy_cor}) is not applied in this problem as the process is isothermal. A four-node hidden layer is used, with all activations comprised of three-point grids. The multiplication hyperparameter $n^{mu}$ in the LeanKAN layer is set to 2 as per the standard LeanKAN formulation~\cite{koenig_leankan_2025}. The training and testing data are sampled from the ground truth at a sparse time interval of 1 s, leading to 30 total data points for each individual species. The ChemKAN is trained on this data for $10^4$ epochs. The leftmost column of Figure~\ref{fig:performance} shows example time-history species profiles for an unseen test case with the ground truth, unseen test data, and prediction by the learned ChemKAN model. With clean, noise-free data, the ChemKAN successfully learns the underlying model and shows good generalizability, with accurate predictions at the training time steps and smooth profiles in between.

A common characteristic of experimental data used for machine learning model inference is the presence of uncertainty or noise, which can cause even well-parameterized deep learning models to overfit as they struggle to distinguish genuine underlying trends from experimental artifacts. This is no different in traditional KANs, where small amounts of noise have been shown to severely limit inference capabilities~\cite{shen_reduced_2024}.
 In the current work, we further probe whether coupling to inherently noise-robust ODE solvers helps ChemKANs to extract useful models from increasingly noisy data. To do so, we task ChemKANs with extracting models from the same dataset with varying amounts of noise added (up to 15\%) as shown in Fig.~\ref{fig:performance}. Surprisingly, the ChemKAN, even with significant amounts of noise, demonstrates strong robustness and a capability to infer smooth and accurate solution profiles that correspond well to the underlying true data. We provide detailed discussions on how ChemKAN performs compared to DeepONet in the following subsections.

\subsubsection{Neural scaling with noise-free data} \label{sec:bio_neural_scaling}

\begin{figure*}[tb]
    \centering
    \includegraphics[width=0.8\linewidth]{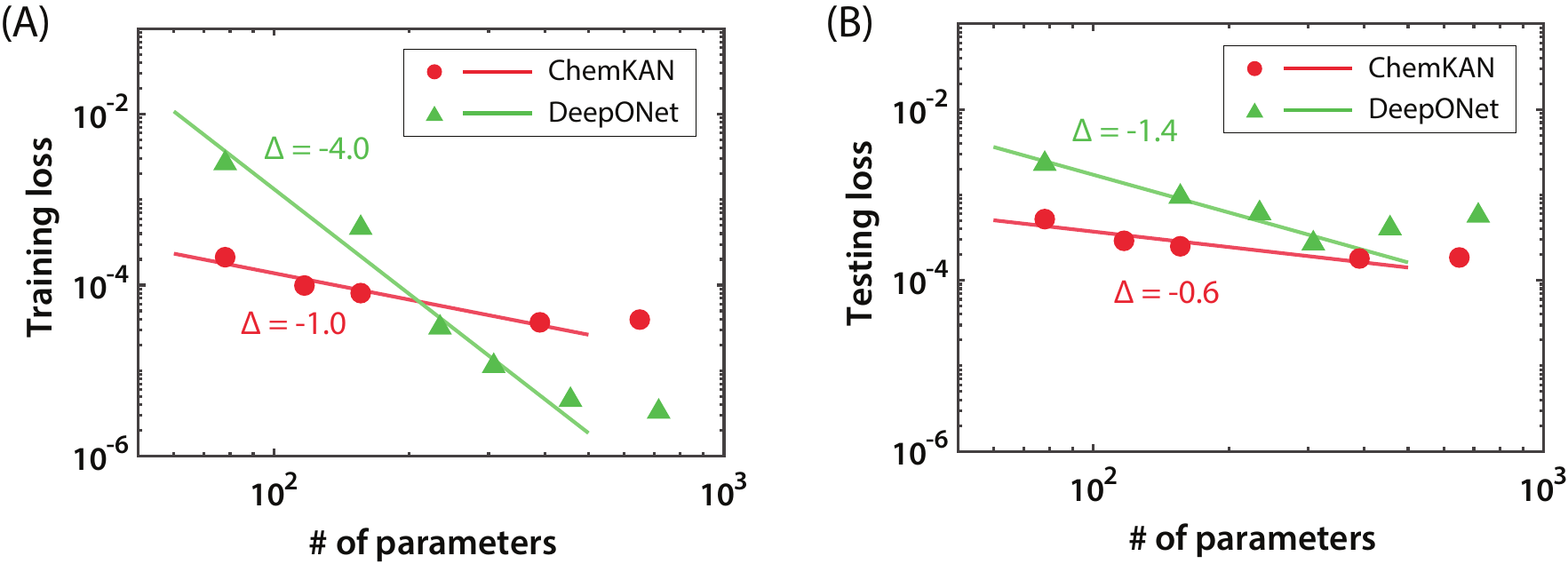}
	\caption{Neural convergence comparison between ChemKAN and DeepONet (no noise). (A) Training MSE results with varying ChemKAN and DeepONet sizes. (B) Testing MSE results with varying ChemKAN and DeepONet sizes. Fitted slopes $\Delta$ in the log scale are provided for each convergence test, where linear fits are evaluated prior to saturation (or in the DeepONet testing results, prior to overfitting).}
	\label{fig:bio_neural}
\end{figure*}

Neural scaling is an effective method of measuring parameter efficiencies in neural networks. Here, different parameter sizes were investigated by changing the number of nodes in the hidden layers. Training and test losses were evaluated for data without added noise, to isolate the underlying expressive capabilities of each technique. Figure~\ref{fig:bio_neural} reveals neural scaling at orders of 1.0 and 0.6 for the training and testing metrics of ChemKANs (where the order of neural scaling is defined as the power to which the loss decreases with respect to the number of parameters). While we might expect values up to 4 based on prior KAN \cite{liu_kan_2024} and KAN-ODE \cite{koenig_kan-odes_2024} studies, we remark here that the low-order neural scaling appears to indicate relatively saturated training of the ChemKAN with errors already around $10^{-4}$ with just 72 parameters, rather than poor convergence which might otherwise have been indicated if low convergence rates were coupled with poor loss metrics.

To further probe the nuances of ChemKAN's convergence efficiency, we compare its results with those of DeepONet. The ChemKAN iterates roughly an order of magnitude slower but converges in fewer epochs (as originally discussed in \cite{koenig_kan-odes_2024}). Thus, to facilitate fair comparison the ChemKAN was trained only for 5,000 epochs, while the DeepONet was trained for 50,000 epochs. Two key distinctions between ChemKAN and DeepONet are as follows. First, as discussed earlier, the extremely sparse ChemKANs (toward the left half of Fig.~\ref{fig:bio_neural}(A)) are able to reach remarkably low error even with just 78 parameters, while the DeepONets see significantly worse performance at sparse parameterizations (also seen in the left half of Fig.~\ref{fig:bio_neural}(A)). Secondly, we note that the DeepONet sees significantly higher order neural convergence in the training results, allowing it to surpass the training performance of the ChemKAN at above 200 parameters, with remarkably strong training accuracy for the largest, 456-parameter DeepONet studied here. Linear fits with slopes are shown in Fig.~\ref{fig:bio_neural}(A) to illustrate this point, where the last ChemKAN and last two DeepONet points are excluded as they begin to plateau in training loss. From this observation, a large-enough DeepONet seems to outperform ChemKAN when looking only at the training losses.

To further contextualize the structural efficiency, we discuss results for the testing error of the same neural convergence runs in Fig.~\ref{fig:bio_neural}(B). Here, we notice a significant departure from the training convergence rates in Fig.~\ref{fig:bio_neural}(A). Unlike the neural convergence for training data, the ChemKAN is seen to outperform the DeepONet at all sizes, with the largest ChemKAN leveling out and retaining nearly the same testing performance as the second-largest ChemKAN. The DeepONet, while enjoying a faster neural convergence rate below 308 parameters, notably fails to plateau and instead appears to diverge at higher parameter counts. When compared against the training results in Fig.~\ref{fig:bio_neural}(A), we observe two distinct modes of training saturation. Saturation, or the point where the linear fit no longer holds, appears to occur for the two largest DeepONets. For the ChemKAN, we might either interpret the entire profile to be saturated, or highlight the single largest network as the saturation point. Regardless, what we observe in these high-parameter networks is high robustness in the ChemKAN to overfitting (with a flat testing loss plateau), compared to the significant overfitting and divergence seen in the DeepONet past 308 parameters (i.e., the last two testing points seeing increasing loss).

It is unsurprising that a standard deep learning technique begins to overfit a small dataset when given a large number of parameters. The DeepONet overfitting past saturation leads to further decreases in training loss accompanied by significant increases in testing loss. What we do find surprising, however, is ChemKAN's apparent resilience to overfitting, even with similar parameter counts. In this context, its original failure to reach the same training performance as the DeepONet appears to be a strength of the method rather than a drawback, as it reaches a minimum value for both training and testing and then remains robust to superfluously added parameters, while the DeepONet clearly requires additional care to avoid overfitting. This pilot neural convergence test suggests that ChemKANs are robust to overfitting, and motivates further study of their capability in a second, more realistic model inference scenario.

\subsubsection{Benefits of latent-dynamics modeling: noisy inference comparison}

\begin{figure*}[tb!]
    \centering
    \includegraphics[width=0.75\linewidth]{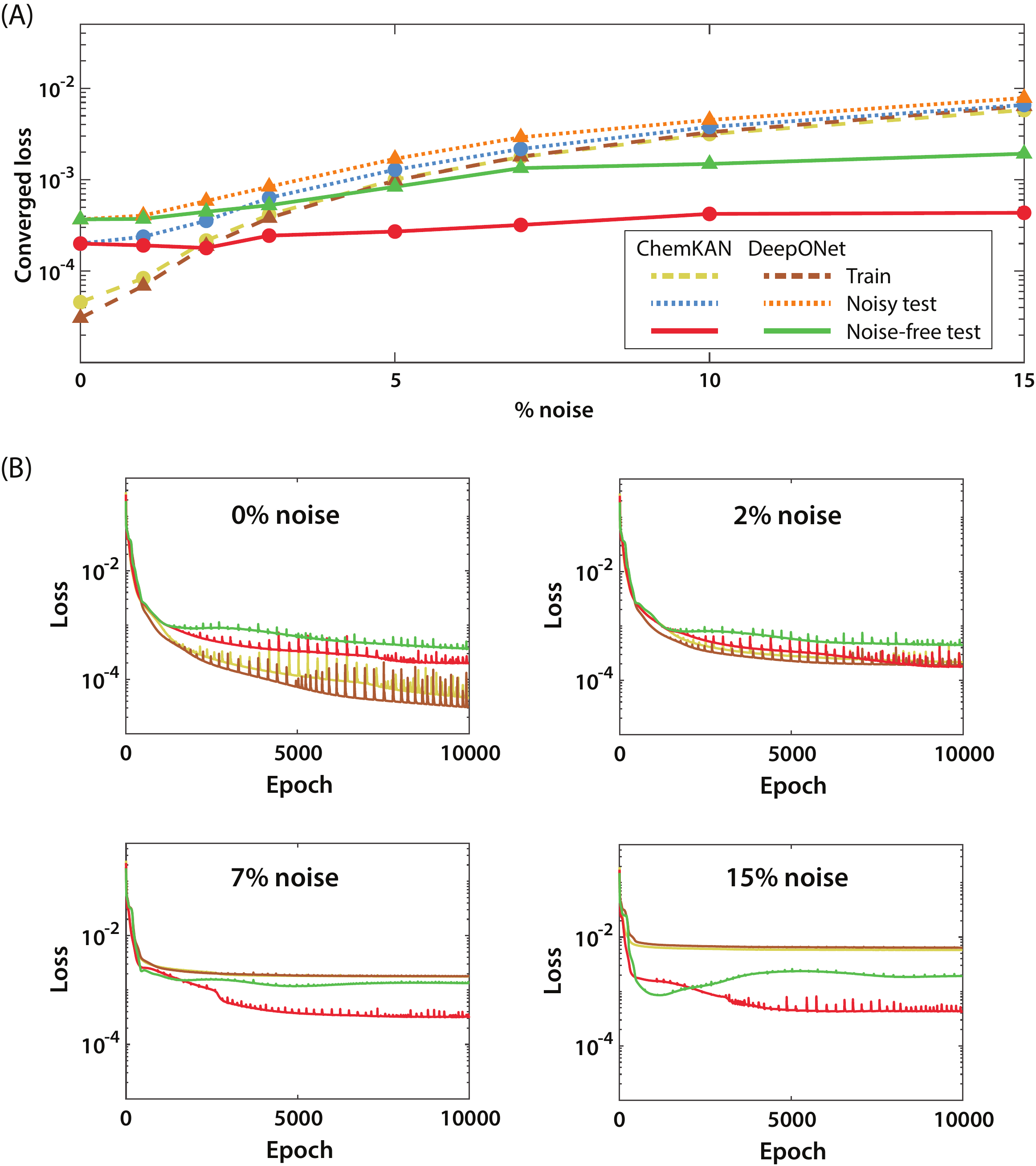}
	\caption{ChemKAN and DeepONet training results with increasing amounts of synthetic noise added to the training data. (A) Converged training, testing, and noise-free testing loss comparisons as a function of \% noise. (B) Comparison of training and noise-free testing loss profiles at increasing amounts of noise (0\%, 2\%, 7\%, and 15\%, respectively). Color legends in (B) are the same with (A). DeepONet is seen to overfit substantially with noisy data, while ChemKAN continues to converge to lower values in all metrics, with no indication of overfitting.}
	\label{fig:bio_training}
\end{figure*}

To gain a deeper understanding of how ChemKAN outperforms DeepONet, we further test the training behaviors of ChemKAN and DeepONet with noisy data. Early stopping is carried out for both network structures at 10,000 epochs, to enable fair comparison and limit overfitting in the DeepONet cases. In all cases, a DeepONet with 308 parameters was used to compare against a ChemKAN with 156 parameters. The 308-parameter DeepONet was chosen based on the results of Fig. \ref{fig:bio_neural}, where this was seen to be the largest DeepONet before the test loss began overfitting. The 156-parameter ChemKAN, meanwhile, was chosen to roughly match the DeepONet's training and testing performance at zero noise. In other words, we chose the best-performing DeepONet size possible based on the preliminary neural convergence study, and then sized the ChemKAN according to the zero-noise performance of both networks. In more detail, the DeepONet had a three-layer branch network with eight nodes per layer and a two-layer trunk network with seven nodes in the first layer and eight in the second layer, with a final output layer converting these two eight-dimensional layers to the six-dimensional solution vector. The ChemKAN, meanwhile, had a single hidden layer with four nodes, two of which included multiplication operators ($n^{mu}=2$) as per the standard LeanKAN formulation \cite{koenig_leankan_2025}, and three gridpoints per activation.

Figure~\ref{fig:bio_training}(A) shows average training results after $10^4$ epochs across the 20 training and 10 testing cases at varied noise levels from $0\%$ to $15\%$. As expected from Sec.~\ref{sec:bio_neural_scaling}, the training MSE with 0\% noise shows that the 308-parameter DeepONet slightly beats the 156-parameter ChemKAN in training performance, and is slightly worse in terms of reconstructing the unseen testing data.

As increasing noise is added to the system, we see in the standard training and testing metrics of Fig. \ref{fig:bio_training}(A) that both networks unsurprisingly see increases in training and testing errors. Roughly, the increase in MSE error in the training losses scales with the square of the noise, as follows from Eq. \ref{eq:loss_h2}. For example, the ChemKAN sees an increase in MSE between 0\% noise and 1\% noise of $3.78 \times 10^{-5}$, where the second degree scaling of the MSE suggests that a 25$\times$ larger increase of $9.45 \times 10^{-4}$ might be expected between 0\% noise and 5\% noise. This is indeed observed, with an increase in this latter case of $9.64 \times 10^{-4} \approx 9.45 \times 10^{-4}$. Thus, the increase in training error for both networks as noise is added can be attributed to the effect of the noise itself on the loss function (Eq.~\ref{eq:loss_h2}), and does not appear to indicate any problems with the two networks' capabilities to fit the increasingly noisy training data. For a direct comparison, The ChemKAN retains lower testing error throughout all tested noise values, and at $7\%$ noise and above, it is actually able to reach a lower training error than the DeepOnet. Looking at the big picture, however, results in these two metrics remain within a factor of two of each other at all noise levels, indicating largely similar performance.

Upon further evaluation, we found that the training and testing MSEs evaluated on noisy datasets do not fully capture the effects of noise on useful model predictions, as overfitting can occur not only to the training conditions but also to the noise present in the data. To more effectively compare these frameworks, we introduce a noise-free MSE metric as was previously studied in the context of KANs~\cite{shen_reduced_2024},

\begin{equation}
        \mathcal{L}_{\text{MSE,noise-free}} = \frac{1}{n^*}\sum_{j=1}^{N_t}\sum_{k=1}^{n^*} \left({\hat{u}^{\text{pred}}_{k}\left( t_{j}, \bm{\theta}\right))- }\hat{u}^{\text{true}}_{k}\left(t_{j}\right)\right) ^{2},
\end{equation}

\noindent which differs from the test MSE $\mathcal{L}_{\text{MSE}}$ in Eq.~\ref{eq:loss_h2} in its use of the true, noise-free data rather than the noisy observations. This noise-free metric serves to quantify the capability of the two modeling approaches to accurately extract the true underlying model from noisy data, rather than overfit the noise or otherwise fail to deliver a useful model. With this metric, much more significant performance shifts can be seen in the noise-free testing values in Fig. \ref{fig:bio_training}(A). The impact of the added noise on the noise-free testing error for the ChemKAN is relatively small (a roughly $2\times$ increase from $0\%$ to $15\%$ noise), suggesting that the ChemKAN is able to extract the true underlying behavior well from the noisy datasets. We reiterate the significance of this result in the context of prior work~\cite{shen_reduced_2024}, where standard KANs were demonstrated to fail when faced with small added noise. In contrast, the DeepONet sees a 5$\times$ increase in noise-free testing MSE from $0\%$ to $15\%$ noise, with its final noise-free testing loss 4.4$\times$ larger than that of the ChemKAN.

\begin{figure*}[tb]
    \centering
    \includegraphics[width=0.95\linewidth]{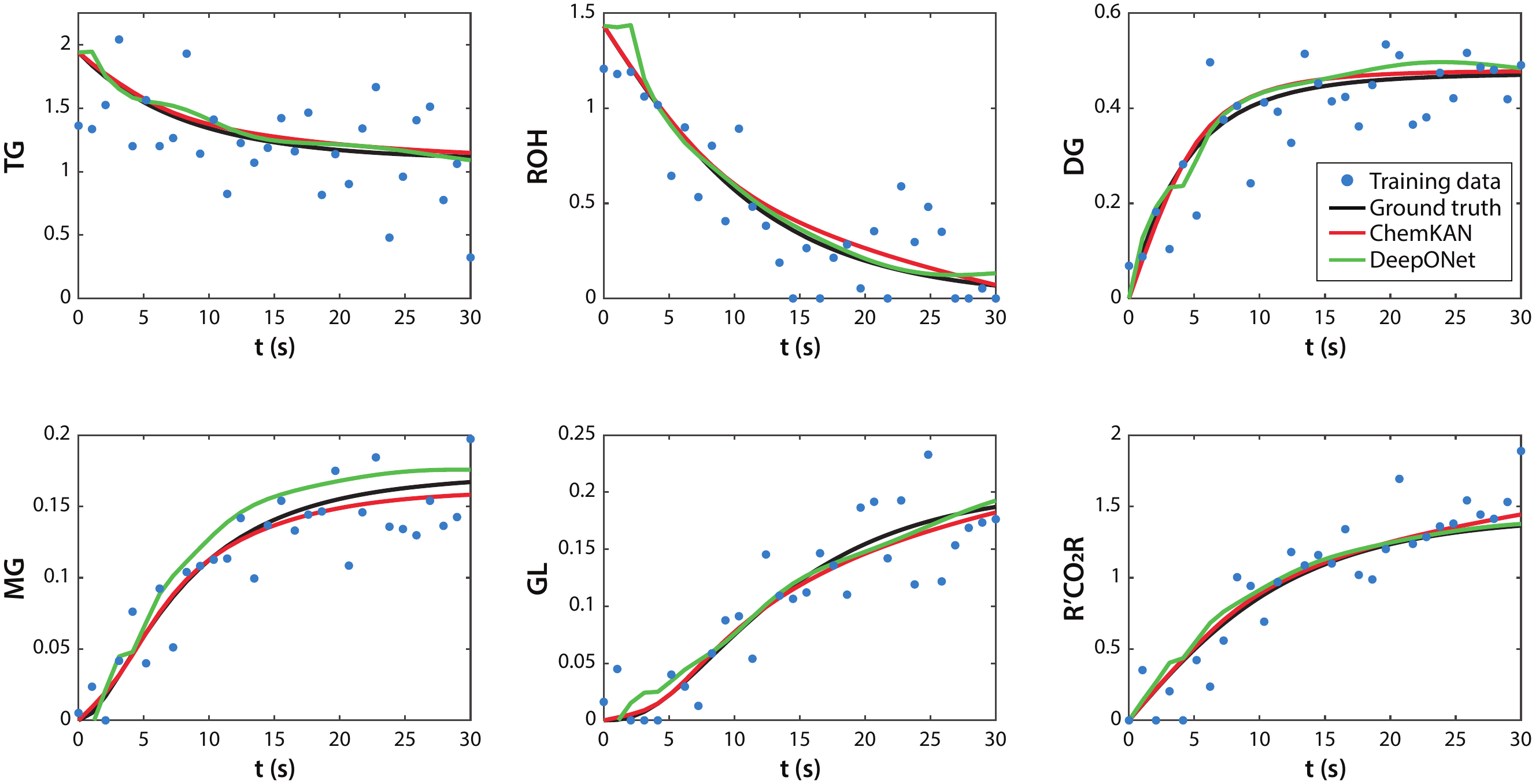}
	\caption{ChemKAN and DeepONet model results trained with 15\% noise, compared against both the noisy training data (blue dots) and the unseen noise-free underlying data (ground truth, black curve). Red and green curves denote ChemKAN and DeepONet, respectively. Where DeepONet exhibits significant overfitting to the training data, ChemKAN is able to more effectively match the unseen ground truth with smoother trajectories.}
	\label{fig:bio_recon}
\end{figure*}

The loss profiles in Figs.~\ref{fig:bio_training}(B) provide further insight on the training dynamics that lead to this significant discrepancy in noise-free testing reconstructions. In the $0\%$ noise training cycle of Fig.~\ref{fig:bio_training}(B), we see fairly standard behavior across the training and noise-free testing traces for the DeepONet and ChemKAN, with all quantities steadily decreasing for the entire duration of training. This is the expected result, as we have sized the DeepONet based on Fig.~\ref{fig:bio_neural} to avoid any overfitting in the noise-free case. As we increase the amount of noise to $2\%$, the training loss values are heavily penalized (due to the noisy data used in the computation of Eq.~\ref{eq:loss_h2}), while the noise-free testing values are slightly penalized but remain comparatively strong, indicating that both approaches are at least to a certain extent able to extract the true underlying model from the noisy data. While the training loss continues to drop quickly and then plateau in all four subplots, we see in the 7\% noise case of Fig.~\ref{fig:bio_training}(B) that the DeepONet noise-free testing loss dynamics begin to suffer, with a minimum value near 5,000 epochs and a slight upward trend toward later epochs, likely due to overfitting. In the 15\% noise case of Fig.~\ref{fig:bio_training}(B), this issue is further exacerbated, with an early minimum near 1,000 epochs followed by significant overfitting to the noisy data for the remainder of the training profile. The ChemKAN in both cases remarkably continues to drop its noise-free testing loss even while the DeepONet is overfitting, with late-epoch dynamics showing plateaued minimum values rather than the overfitting seen in the DeepONet. This echoes the behavior seen in Fig.~\ref{fig:bio_neural}, where the ChemKAN did not overfit and instead simply plateaued at its minimum training and testing errors.

Reconstructed training data profiles are shown for the 15\% noise case in Fig.~\ref{fig:bio_recon}. While the DeepONet and ChemKAN are both roughly able to find the unseen, noise-masked ground truth profile, a close inspection reveals not only better ChemKAN fits but also notably jagged profiles from the DeepONet as it attempts to overfit the noise present in the training data. These results suggest that the dynamical system exploitation inherent to the ODE-based framework of ChemKANs (and KAN-ODEs in general) can mitigate or even entirely resolve previously observed issues~\cite{shen_reduced_2024} regarding noisy data with KANs, and help to surpass the performance of standard DeepONets.

In summary, we have demonstrated ChemKANs as a promising tool for model discovery in temperature-dependent chemical kinetic systems, especially with realistically noisy datasets. They show promise not only on the scientific side of the problem, where they were demonstrated here to have significant capability compared to a standard tool in discovering models hidden under noisy data, but also on the machine learning side of the problem, where we have demonstrated that the neural ODE implementation of KAN-ODEs and ChemKANs helps them to overcome the noisy data limitation recently shown in vanilla KANs \cite{shen_reduced_2024}. 

\subsection{Hydrogen combustion acceleration from homogeneous reactor data} \label{results:H2}

\begin{figure*}[!htb]
    \centering
	\includegraphics[width=0.65\linewidth]{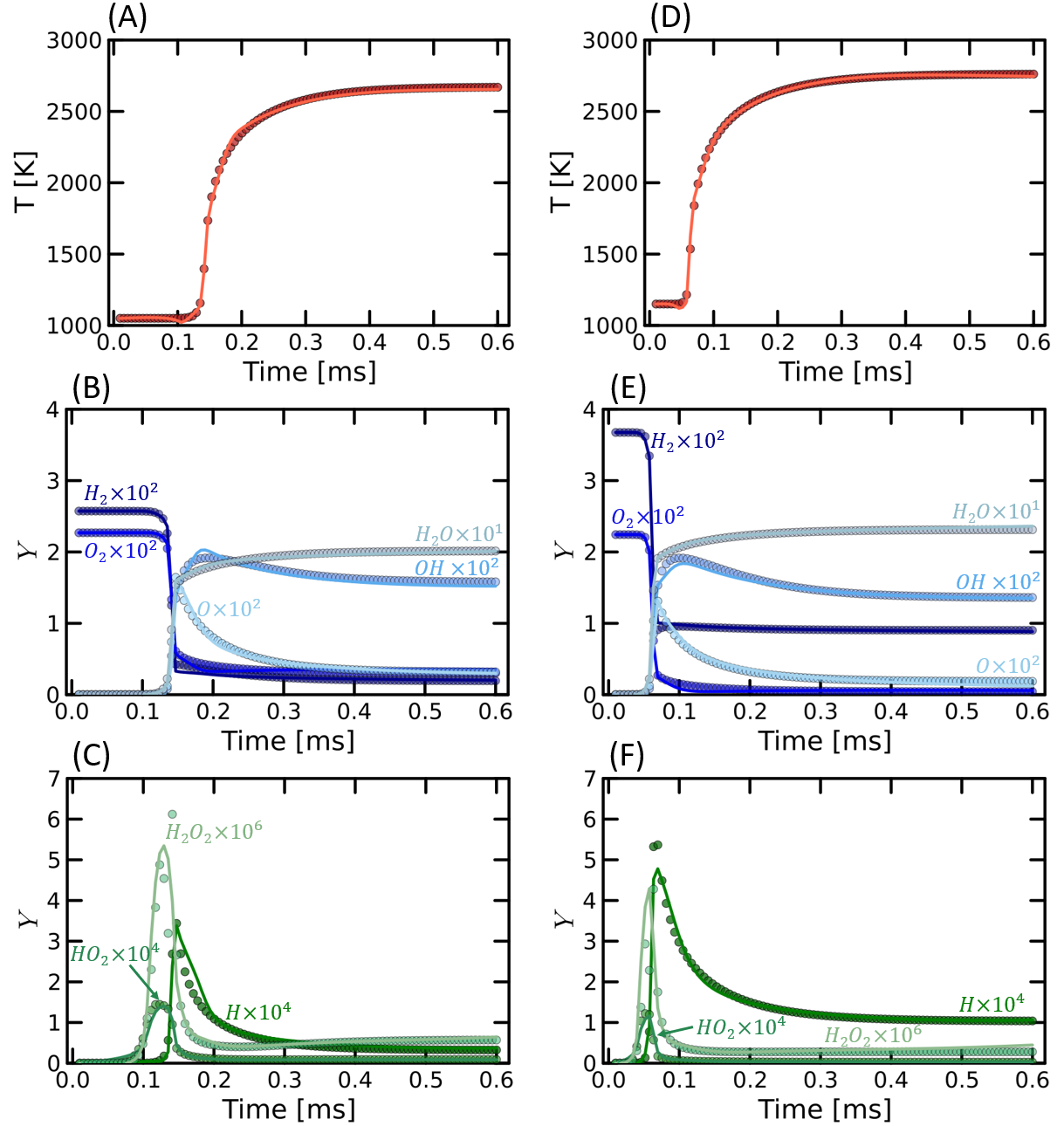}
	\caption{KAN-ODE reconstruction of homogeneous reactor results. (A, B, C) temperature, species reconstructed here that were originally studied in ChemNODE, and additional low-concentraiton species and radicals studied only here, respectively, at a training condition of $\Phi=0.9$ and $T_0=1050$ K. (D, E, F) same three subfigures at the unseen testing condition of $\Phi=1.3$ and $T_0=1150$ K. Dots are the ground truth, and curves are ChemKAN reconstructions. Constant $N_2$ profiles are not plotted to improve clarity for reacting quantities, as training data does not include $NO_x$ chemistry. } 
	\label{fig:H2_results}
\end{figure*}

In our second case study, we investigate the use of ChemKANs as a reduced-order solver acceleration framework for known chemical models. We hypothesize based on previous high-order neural scaling results \cite{koenig_kan-odes_2024, liu_kan_2024} as well as the strong performance at low parameter counts seen in Fig. \ref{fig:bio_neural} that the use of a KAN structure instead of an MLP will allow for similar accuracy in network predictions with fewer parameters and lower cost. We additionally aim to evaluate the novel network architecture of the ChemKAN, and whether its physics-based, single-network structure is able to model behavior that required $m+1$ MLPs in the Neural ODE framework~\cite{owoyele_chemnode_2022}. To reduce computational cost to the furthest extent possible, we use a single hidden layer of just three nodes for the core kinetic network (Eq.~\ref{eq:chemkan_species}). This three-node hidden layer works as a latent representation of the system's dynamics, compressing the information from 10 thermochemical states in the input and 29 reactions in the kinetic model. Leveraging our knowledge of the functional form of the governing equations (Eqs.~\ref{eq:species} and \ref{eq:energy}) and their strong multiplicative behavior, we use $n^{mu}=3$ here, defining all three hidden nodes using the multiplication operator.

We begin in Fig.~\ref{fig:H2_results} with a demonstration of the ChemKAN homogeneous reactor reconstructions for one training case (left column, $\Phi=0.9$ and $T_0=1050$ K) and the unseen testing case (right column, $\Phi=1.3$ and $T_0=1150$ K), after the two-stage training process outlined in Sec.~\ref{methods:h2}. These reconstructions were generated entirely by the ChemKAN, given only the initial conditions. Overall the learned model successfully predicted the temperature and mass fractions in both cases, with no notable deterioration in the testing case (as expected from the strong testing results and robustness to overfitting observed in the previous biodiesel investigation). We further emphasize that the ChemKAN was largely able to capture the behavior of the low-concentration and highly reactive species $Y_{\text{H}_{2}\text{O}_{2}}$, $Y_{\text{H}}$, and $Y_{\text{HO}_{2}}$ (Fig.~\ref{fig:H2_results}(C, F)) that were neglected in ChemNODE \cite{owoyele_chemnode_2022}.

\begin{figure*}[tb!]
    \centering
	\includegraphics[width=0.65\linewidth]{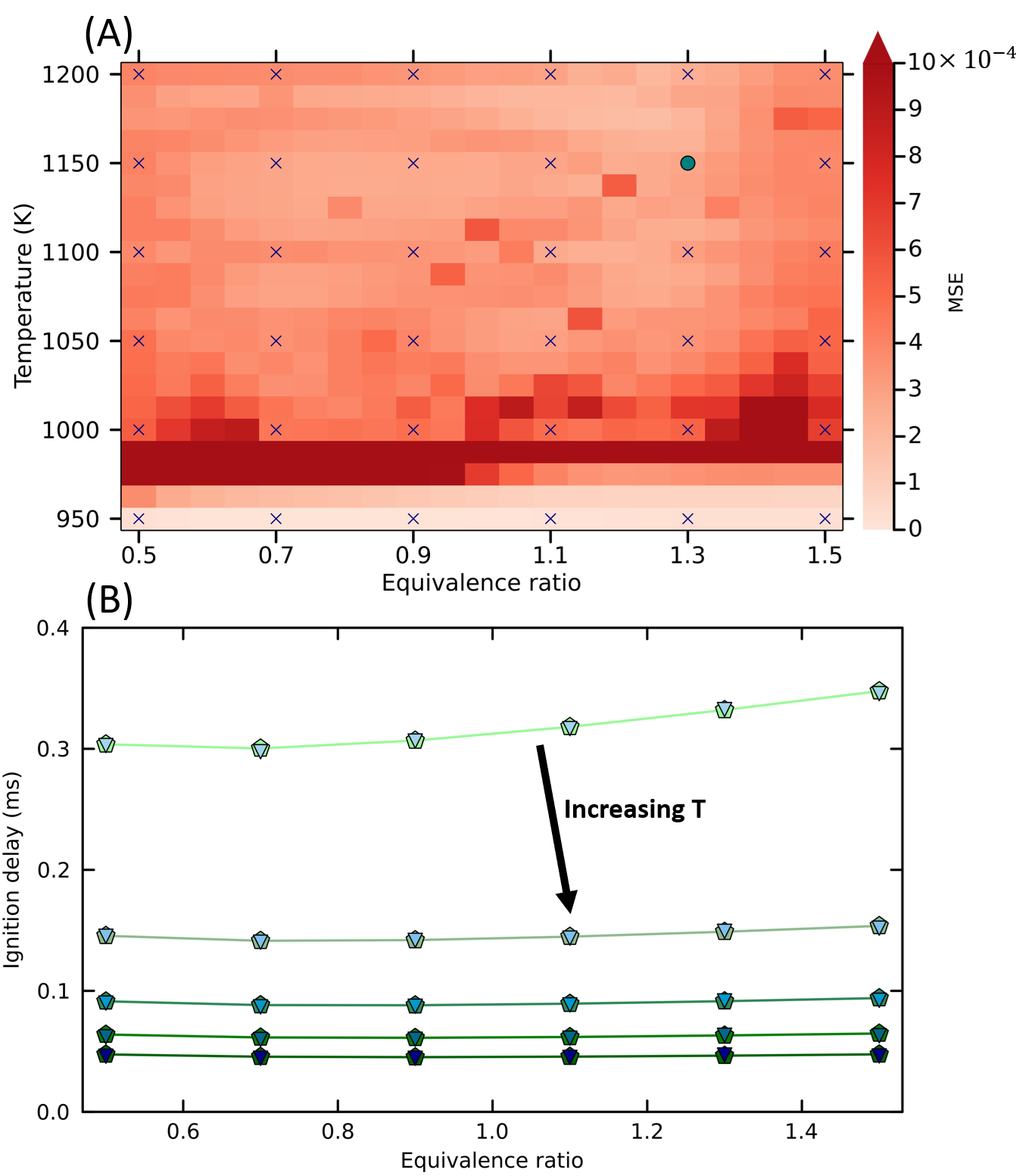}
	\caption{Evaluation of the proposed ChemKAN framework for various conditions. (A) ChemKAN  reconstruction error at 35 training initial conditions (navy crosses), single testing initial condition plotted in Fig. \ref{fig:H2_results} (teal dot), and 405 additional testing locations between the initial 36.  (B) Actual and ChemKAN-predicted ignition delay times (green pentagons and blue triangles, respectively), as a function of equivalence ratio, at different initial temperatures (1000 K, 1050 K, 1100 K, 1150 K, and 1200K, from top to bottom).}
	\label{fig:heat_and_idt}
\end{figure*}

A broader comparison is shown in Fig.~\ref{fig:heat_and_idt}(A), where the MSE (in the normalized $\textbf{u}$ units) is plotted across not only the initial set of 35 training and 1 testing initial conditions, but a wider set of 441 total initial conditions (406 of which were unseen during training) at a finer resolution in the same range. The low-temperature initial conditions see near-perfect reconstructions, as the ignition delays there are larger than the studied time window, leading to smooth gradients and easily trainable, near-isothermal behavior. In the remainder of the domain, strong performance is seen at all training conditions, and additionally at the single testing condition plotted in Fig. \ref{fig:H2_results}. In terms of generalization to intermediate temperatures and equivalence ratios, the ChemKAN performed very well throughout the vast majority of the domain, with many testing points even surpassing the accuracy of nearby training points at and above 1050 K. We notice that toward the slower-igniting cases, however, the ChemKAN struggled more with generalization. At 1000 K, for example, all six training points saw strong MSE values in the $10^{-4}$ range, although a few of the intermediate equivalence ratios suffered. At 987.5 K, however (one tick below 1000 K), all testing points saw poorer performance in the $10^{-3}$ range. We can conclude from these results that the ChemKAN retains its strong capability to generalize in the more challenging H$_{2}$-air combustion case (as was originally reported in the biodiesel modeling case), but with practical limits \hl{in the colder, more temperature-sensitive ignition cases. While the ChemKAN's ability to accurately learn the six studied training points at 1000 K suggests that it is capable of tracking ignition behavior through the cooler regions, its relatively poor performance elsewhere in the initiation-sensitive regime suggests that a non-uniform training grid with denser sampling toward such cooler regions is needed to fully resolve this behavior and provide more accurate results when applied in combustion CFD simulations that require accurate ignition behavior.}

We finally plot the actual and KAN-ODE predicted ignition delays in Fig. \ref{fig:heat_and_idt}(B), for the 30 studied cases that ignited (the lowest temperature cases did not see ignition given the time span of 0.6 ms). Ignition here is defined as the point of maximum temperature rise rate~\cite{owoyele_chemnode_2022}. Accuracy is strong across the board, even in the testing case.

This collection of results shows that the ChemKAN structure was able to accurately learn the dynamics of all nine species and temperature scalars across the same set of initial reactor conditions as was studied using traditional MLP-based Neural ODEs in ChemNODE \cite{owoyele_chemnode_2022}. Compared to the six species plus temperature scalars learned there via seven unique MLP networks with 91 parameters each (according to standard MLP parameterizations, 637 total parameters), the current ChemKAN was able to learn the complete set of thermochemical scalars (nine species plus temperature) using a single, 344-parameter network. While training took place in two stages to decouple the kinetic and thermodynamic behavior and facilitate convergence, the final network remains a single cohesive structure with shared information across all nodes, eliminating the redundancies in repeated yet isolated 91-parameter MLPs. 

Finally, regarding computational efficiency, we report that the average time to solve all 36 homogeneous reactor conditions in the Arrhenius.jl combustion solver package \cite{ji_arrheniusjl_2021} was 2$\times$ faster when switching from the detailed chemistry to the reduced ChemKAN framework. The total number of time steps in the integration process remains largely unchanged, as the ChemKAN is solving the same full-dimensional thermochemical state. The 2$\times$ speedup is predominantly achieved through faster gradient computation (i.e., each time step is faster to compute), facilitated by the ChemKAN's compression of 29 reactions into just three hidden nodes and a handful of sparse activations. By learning the relationship between the current thermochemical state and the chemical source terms, the ChemKAN is capable not only of predicting ignition delay times and homogeneous reactor solution profiles 2$\times$ faster than the detailed model, but also of generalizing to other simulation conditions\hl{ when coupled to flow solvers,} including simple laminar flames and complex 2-D and 3-D turbulent combustion conditions. Such downstream uses of similar surrogate machine learning models were discussed and tested in previously~\cite{jung_hessian-based_2024, owoyele_chemnode_2022}, where a 2$\times$ speedup in the chemical solver (which is often the most computationally expensive component in a reacting flow simulation) implies the potential for substantial acceleration unlocked by ChemKANs while retaining the full-sized, detailed solution state vector. While slightly slower than the 2.3$\times$ speedup reported in ChemNODE \cite{owoyele_chemnode_2022}, we reiterate that the ChemKAN solves for an additional three minor species (including the key H radical). A summarized comparison of ChemKAN and ChemNODE is provided in Table~\ref{table:comparison}.

\begin{table}
    \centering
    \caption{Efficiency comparison between ChemNODE \cite{owoyele_chemnode_2022} and ChemKAN. Note that a ChemNODE tracking all ten thermochemical quantities would require 1210 parameters, as a single network in this case would comprise 121 parameters. Species in bold indicate missing species in ChemNODEs.}
    \vspace{5pt}
    \begin{tabular}{p{0.2\linewidth} p{0.07\linewidth} p{0.09\linewidth} p{0.21\linewidth} p{0.145\linewidth}}
    \toprule
    & \# of nets & \# of params & Species modeled & Speed-up vs. true model \\
    \midrule
    ChemNODE\cite{owoyele_chemnode_2022} & 7 & 637 & H$_2$, O$_2$, H$_2$O, N$_2$, O, OH & 2.3$\times$ \\
    \midrule 
    ChemKAN (our work) & 1 & 344 & H$_2$, O$_2$, H$_2$O, N$_2$, O, OH, \textbf{H}, \textbf{HO$_2$}, \textbf{H$_2$O$_2$} & 2.0$\times$ \\ \bottomrule
    \end{tabular}
    \label{table:comparison}
\end{table}

\subsection{Current limitations and areas for further research}

While we have demonstrated that our ChemKANs provide remarkable inference capability and expressivity, there remain drawbacks. The total speedup reported in Table~\ref{table:comparison} is ultimately not as large as we believe that it could be. While still more than competitive with that of ChemNODE, especially considering the complete thermochemical source term with low-concentration radicals that ChemKAN provides, it appears underwhelming in light of the rest of the significant performance gains enabled by ChemKAN in other comparisons throughout this work. That being said, we believe that the reported speedup is a conservative lower bound on the potential for ChemKAN and KANs in general. The original KANs \cite{liu_kan_2024} are, at the time of original submission of the current work, barely a year old and are known in the literature to be much slower than comparable MLPs (this is even acknowledged by the authors of the original KAN paper \cite{liu_kan_2024}). However, extensive research is actively ongoing to resolve this issue, including parameter efficiency improvements \cite{koenig_leankan_2025, lee_hippo-kan_2024, ta_af-kan_2025} and prediction acceleration techniques \cite{blealtan_efficient-kan_2024, chen_lss-skan_2024, huang_hardware_2025, moradzadeh_ukan_2024, qiu_relu-kan_2024, puri_kolmogorovarnoldjl_2024, qiu_powermlp_2024}. It is unclear which methods will ultimately prevail, but the recent emergence of KANs and the large amount of work proposing various techniques for their acceleration suggest significant promise in the near future for substantial and relatively lower-effort acceleration, compared to the more mature and well-developed MLPs where we believe it reasonable to expect a slower pace of future development. 

We have additionally in this section compared a baseline ChemKAN implementation against a baseline ChemNODE implementation. Later works exist that appear to successfully combine ChemNODE with augmented loss functions, autoencoders, and latent space time stepping. The most recent, ``Phy-ChemNODE'', includes all of these techniques~\cite{kumar_physics-informed_2024}. We do not draw comparisons between the current ChemKAN implementation and the larger-scale, combined-methodology results reported there, as our current aim is to compare the pure performance of ChemKAN against the MLPs that underlie both ChemNODEs and Phy-ChemNODEs. All further augmentations carried out in Phy-ChemNODE that go beyond this baseline can be replicated with ChemKANs, and an interesting target of future studies may be to quantify the performance gains of ChemKAN when applied in tandem with other advanced neural network structures.

\section*{Conclusions}

This work introduced ChemKANs, a novel physics-informed machine learning technique based on the general KAN-ODE framework with a specialized structure tailored explicitly for chemical kinetic modeling and acceleration. Its two-part design allows for application in isothermal or exothermic systems through a kinetic core and an optional thermodynamic superstructure that can be trained and applied as a single cohesive framework. Model inference in a preliminary biodiesel synthesis case using only the kinetic core revealed that ChemKANs are remarkably robust to overfitting. Fair comparisons against a generic DeepONet approach revealed that with increasingly bulky parameterizations, ChemKANs retained plateaued optimal loss metrics, while DeepONets saw minor further decreases in training loss accompanied by an increase in testing loss, as might be expected from a standard deep learning approach. With added noise, this difference was further exacerbated, with the DeepONet achieving low training loss through jagged fits to the noise itself, while the ChemKAN converged its training, testing, and noise-free testing performance with smooth fits to the hidden underlying data. While alone promising, these results are of particular interest given prior works in the literature that have cast doubts on the effectiveness of KAN structures on noisy functions.

In a second case, ChemKANs were demonstrated as efficient acceleration surrogates for learning chemical source terms in a hydrogen combustion case. A two-stage training process for the kinetic core and thermodynamic superstructure enabled a single, 344-parameter ChemKAN to accurately learn complete solution profiles across a range of hydrogen-air homogeneous reactor initial conditions, a significant reduction in parameter and network bulk compared to previous MLP-based neural ODE approaches that required 637 parameters to learn a truncated set of solution profiles. Timing comparisons against the detailed mechanism revealed a 2$\times$ speedup when using the ChemKAN surrogate model, which is significant for downstream applications of the hydrogen combustion surrogate learned here (for example, 3-D turbulent reacting flow). In summary, we find that ChemKANs are a promising tool for both dynamical system modeling and acceleration tasks in combustion chemistry. In doing so, we have also successfully advanced the underlying KAN-ODE framework to much larger, practical systems than had been studied previously. We hope that these promising preliminary case studies motivate future implementation of ChemKAN layers and modules in combustion and chemical kinetic machine learning applications.

\section*{Author contributions}
\textbf{Benjamin C. Koenig:} Conceptualization, Methodology, Software, Investigation, Writing - Original Draft. \textbf{Suyong Kim:} Conceptualization, Methodology, Writing - Original Draft, Writing - Review $\&$ Editing. \textbf{Sili Deng:} Funding Acquisition, Resources, Writing - Review $\&$ Editing.

\section*{Conflicts of interest}
There are no conflicts to declare.

\section*{Data Availability}

The data used to produce the results in this manuscript are available on request.

\section*{Acknowledgements}

This work is supported by the National Science Foundation (NSF) under Grant No. CBET-2143625. BCK is partially supported by the NSF Graduate Research Fellowship under Grant No. 1745302. 


\bibliography{ChemKAN} 
\bibliographystyle{elsarticle-num}

\end{document}